\documentclass[letterpaper, 10 pt, conference]{IEEEtran}  

\IEEEoverridecommandlockouts                              

\usepackage{tabularx}
\usepackage{subcaption}
\usepackage{float}
\usepackage{amsmath}
\usepackage{amssymb}
\usepackage[usenames, dvipsnames, table]{xcolor}
\usepackage{graphicx}

\usepackage[normalem]{ulem}
\usepackage{hyperref}
\usepackage[capitalize]{cleveref}
\usepackage{bm}
\usepackage{multirow}
\usepackage{booktabs}

\usepackage[
    style=ieee,
    sorting=none,
    backend=biber,
    maxbibnames=4,
    doi=true,
    url=false,
    eprint=true,  
    isbn=true
]{biblatex}
\usepackage[capitalize]{cleveref}
\usepackage{algorithm}
\usepackage{algpseudocode}
\usepackage{xcolor}
\usepackage{makecell}

\let\labelindent\relax
\usepackage[inline]{enumitem}

\colorlet{mylinkcolor}{BrickRed}
\colorlet{mycitecolor}{Green}
\colorlet{myurlcolor}{NavyBlue}
\hypersetup{
  linkcolor  = mylinkcolor,
  citecolor  = mycitecolor,
  urlcolor   = myurlcolor,
  colorlinks = true,
}

\begin{document}

\title{ 
Physically Accurate Rigid-Body Dynamics in Particle-Based Simulation
}

\author{
Ava Abderezaei*, Nataliya Nechyporenko, Joseph Miceli, Gilberto Briscoe-Martinez, Alessandro Roncone
\thanks{$^{*}$Corresponding author.}%
\thanks{GBM is supported by NASA Space Technology Graduate Research Opportunity Grant 80NSSC22K1211.}%
\thanks{All Authors with Department of Computer Science, University of Colorado Boulder, CO, USA.}%
\thanks{Emails: {\tt\small \{firstname.lastname\}@colorado.edu}}%
}
\maketitle

\begin{abstract}
Robotics demands simulation that can reason about the diversity of real-world physical interactions, from rigid to deformable objects and fluids. Current simulators address this by stitching together multiple subsolvers for different material types, resulting in a compositional architecture that complicates physical reasoning. Particle-based simulators offer a compelling alternative, representing all materials through a single unified formulation that enables seamless cross-material interactions.
Among particle-based simulators, position-based dynamics (PBD) is a popular solver known for its computational efficiency and visual plausibility.  However, its lack of physical accuracy has limited its adoption in robotics.
To leverage the benefits of particle-based solvers while meeting the physical fidelity demands of robotics, we introduce PBD\nobreakdash-R, a revised PBD formulation that 
enforces physically accurate rigid-body dynamics through a novel momentum-conservation constraint and a modified velocity update.
Additionally, we introduce a solver-agnostic benchmark with analytical solutions to evaluate physical accuracy. Using this benchmark, we show that PBD\nobreakdash-R significantly outperforms PBD and achieves competitive accuracy with MuJoCo while requiring less computation.

\end{abstract}

\section{Introduction}
\label{sec:intro}

Simulation is central to modern robotics, underpinning policy learning, model-based control, planning, and sim-to-real transfer. Robots rely on simulation to reason about how their actions impact the world: for example, pushing a container of liquid requires anticipating both the container’s motion and the fluid’s dynamics.
However, the typical simulation stack is fragmented in how it models diverse materials. It uses separate engines that are ``best" for specific physical properties: rigid-body engines for rigid objects, finite element methods for deformables, and smoothed particle hydrodynamics for fluids \cite{nvidia2025isaaclabgpuacceleratedsimulation}. 
As a result, simulators have to rely on heterogeneous subsolvers to model complex environments.
This fragmentation has significant downstream consequences.
It forces each material to rely on specialized contact models, complicating cross-material interaction and breaking the end-to-end differentiability required for gradient-based optimization and learning. Errors compound across incompatible subsolvers, making simulation pipelines brittle and difficult to scale to the contact-rich manipulation tasks robots must perform in the real world.

\begin{figure}
    \centering
    \includegraphics[width=\columnwidth]{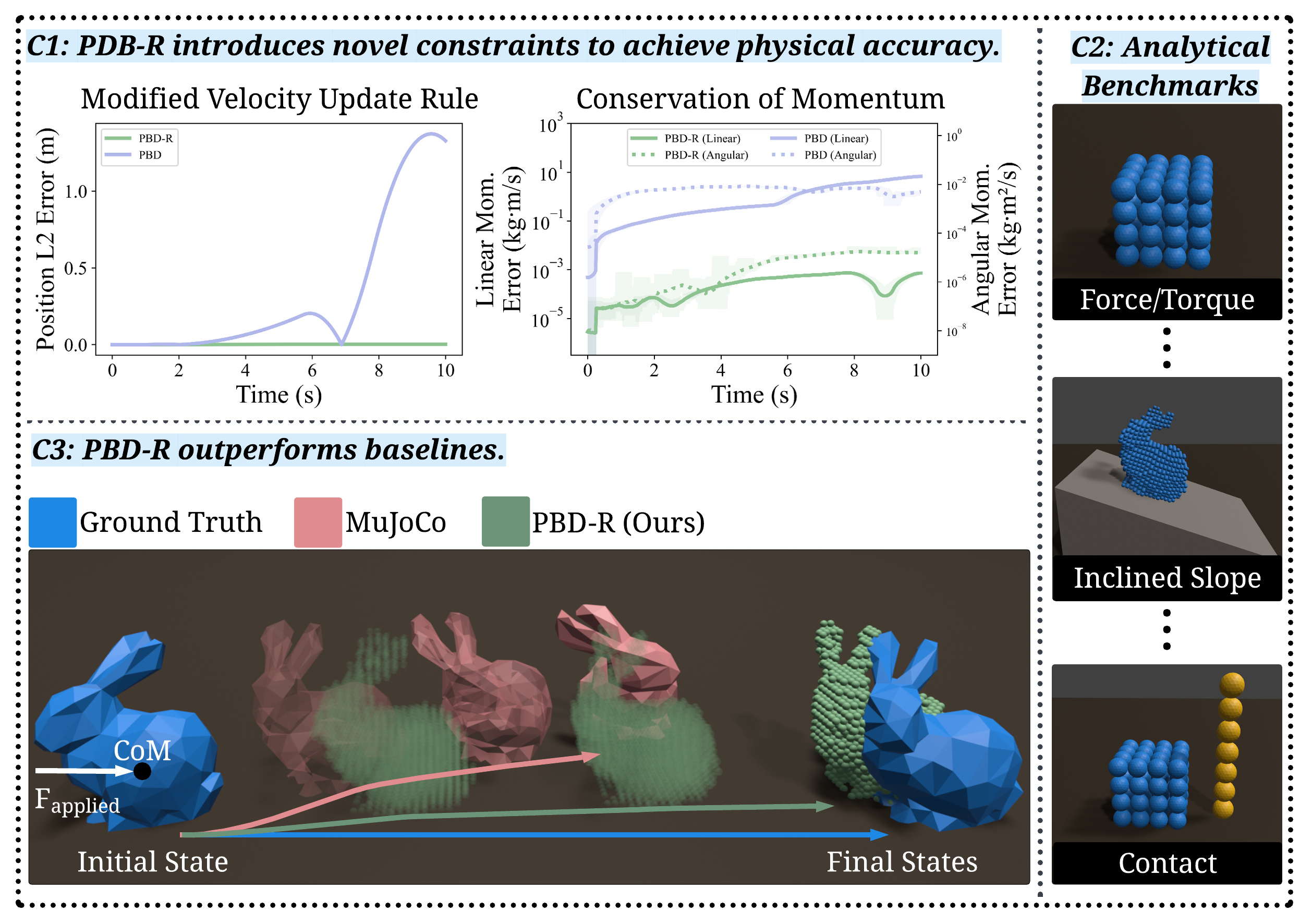}
    \caption{
To leverage the benefits of particle-based simulation for robotics, we introduce PBD-R, a revised Position-Based Dynamics formulation that addresses the rigid-body physical accuracy gap in standard PBD. Contribution 1 (Top): PBD-R introduces modifications to PBD to achieve physical accuracy.
Contribution 2 (Right): We introduce a solver-agnostic benchmark of seven physics tests with analytical solutions for evaluating simulator accuracy. Contribution 3 (Bottom): PBD-R outperforms PBD and achieves competitive accuracy to MuJoCo while requiring less computation.
    }
    \label{fig:first_page}
\vspace{-2em}
\end{figure}

Within this fragmented landscape, particle-based representations have emerged as a compelling alternative. Unlike the mesh-based pipelines that underpin the majority of existing simulators, particle-based methods offer a fundamentally more expressive and unified representation for physical simulation.
By treating rigid, deformable, cloth, and fluid entities as a set of particles, a single contact model and solver can handle all interactions uniformly \cite{10.1145/2601097.2601152}.
Beyond unified simulation, particle-based representations 
provide additional advantages due to their meshless formulation: 
They support direct collision checking without requiring convex decomposition of complex geometry \cite{vu2025empartinteractiveconvexdecomposition}, enable faster collision computation through simplified proximity queries \cite{nechyporenko2025morphit}, and permit straightforward digital twin construction by mapping from point clouds to simulation-ready particle sets \cite{abou-chakra2024physically}. 
Position-Based Dynamics (PBD) is a particularly attractive particle-based framework due to its computational efficiency, numerical stability, and support for a wide range of constraint types~\cite{10.1145/2601097.2601152, 10.1145/2994258.2994272, 10.2312/egt.20171034}. A recent variant of PBD \cite{10.1145/3606923}, further enables differentiable simulation.
It has also seen increasing adoption in robotics for tasks such as cloth manipulation~\cite{zhang2023achievingautonomousclothmanipulation} and digital twin construction~\cite{abou-chakra2024physically}.

However, PBD was originally developed for computer graphics, where computational efficiency and visual plausibility are prioritized over physical accuracy~\cite{10.2312/egt.20171034}. 
This is a critical limitation for robotics, where physical fidelity directly determines the sim-to-real gap~\cite{aljalbout2025reality} and the downstream viability of policy transfer to the real world.
Reducing this gap requires simulators that model physical dynamics with high accuracy. 
To this end, we introduce PBD\nobreakdash-R, which improves the physical accuracy within the unified particle representation paradigm of PBD. PBD\nobreakdash-R improves the physical accuracy of rigid-body dynamics without sacrificing the benefits of the particle-based framework.

To measure PBD-R's improvements, rigorous, quantifiable metrics of physical fidelity are required. However, systematically measuring how accurately a simulator reproduces real-world physics remains an open problem.
To that end, we introduce a physics benchmark consisting of seven tests with their corresponding analytical solutions. This benchmark is hardware-free, solver-agnostic, and easily extensible, enabling systematic comparison of simulation methods.
Moreover, it complements common evaluation practices that focus on task-level metrics such as reward curves or sim-to-real transfer 
\cite{acosta2022validatingroboticssimulatorsrealworld, kaup2024reviewphysicsenginesreinforcement, abderezaei2024clutterawarespillfreeliquidtransport}
and enables direct observation of error by measuring deviation from analytical solutions.
Using this benchmark, we conduct an extensive evaluation of PBD\nobreakdash-R, comparing it against PBD and MuJoCo's \cite{6386109} rigid-body solver.
Our results show that PBD\nobreakdash-R consistently outperforms standard PBD and achieves competitive accuracy to MuJoCo, while requiring fewer computational resources. 


In summary, our contributions are: (1) A revised PBD formulation, PBD\nobreakdash-R, with a novel constraint and velocity update modification that achieves a significantly improved physical accuracy. (2) A solver-agnostic physics benchmark comprising seven tests with analytical reference solutions, providing a general framework for quantitative evaluation of physical accuracy in physics simulators. (3) An extensive empirical evaluation comparing PBD\nobreakdash-R with standard PBD and MuJoCo across all seven benchmark tests.
Collectively, our work paves the way toward a single high-fidelity, fully differentiable, and computationally efficient simulator that accurately captures the complexity of real-world physics.

\section{Related Work}
\label{sec:related}

Modern robotics simulators rely on several physics subsystems to model objects with different material behaviors. For example, MuJoCo’s ~\cite{6386109} core engine simulates articulated rigid bodies with a soft‑constraint, convex‑optimization contact solver. However, deformable objects such as cables, cloth, and soft boxes are modeled as FEM‑style objects with their own contact and material models. PyBullet~\cite{coumans2021} similarly pairs a constraint-based multibody rigid solver with a separate soft-body module that implements cloth and volumetric deformables using FEM, mass–spring, or position-based dynamics backends, while experimental SPH fluid modules exist as yet another particle-based subsystem.  NVIDIA Isaac ~\cite{makoviychuk2021isaacgymhighperformance, nvidia2025isaaclabgpuacceleratedsimulation} builds on PhysX to expose three distinct GPU-accelerated pipelines: a rigid and articulated body solver, a finite-element-based deformable body solver operating on tetrahedral meshes, and a position-based dynamics particle system for fluids, granular media, and cloth.
Relying on fragmented simulation subsystems inherently complicates cross-material dynamics and breaks differentiability, resulting in brittle pipelines that fail to capture the contact-rich interactions required for real-world robotics.

Particle-based simulators present a compelling alternative. A unified particle formulation enables diverse material behaviors to be modeled within a single framework \cite{10.1145/2601097.2601152}. Their mesh-less nature further provides practical advantages, including collision checking without convex decomposition \cite{vu2025empartinteractiveconvexdecomposition}, faster collision computation \cite{nechyporenko2025morphit}, and digital twin construction by mapping point clouds to simulation-ready particle sets \cite{abou-chakra2024physically}.
Among particle-based methods, Position-Based Dynamics is a popular choice in computer graphics due to its computational efficiency, numerical stability, and ability to create visually plausible outputs.
PBD supports a wide range of constraints \cite{10.2312/egt.20171034} that enable it to model diverse material properties (rigid, deformable, fluid, cloth, etc.). Extended Position-Based Dynamics (XPBD) extends PBD to better handle elastic and dissipative energies, improving the stability and realism of deformable objects~\cite{10.1145/2994258.2994272}. More recently, a differentiable variant of PBD has further expanded its capabilities~\cite{10.1145/3606923}. PBD has been explored in robotics as well. For example, \cite{zhang2023achievingautonomousclothmanipulation} creates a quasi-static XPBD formulation for physics-aware cloth manipulation. Work by \cite{abou-chakra2024physically} leverages PBD to construct a digital twin. \cite{abou-chakra2024physically} relies on PBD for its computational efficiency and compensates for physical accuracy through real-to-sim visual force calculation. These works highlight that physical inaccuracy is the primary cause of PBD's sim-to-real gap in robotics, therefore we designed PBD-R to directly address physical fidelity.

To quantify physical fidelity, we create a benchmark of seven solver-agnostic physics tests with known analytical solutions, illustrated in \cref{fig:tests}, and compare our method to an analytical solution. Our benchmark complements existing evaluation practices for simulators, which typically include: comparing simulation results to real-world data \cite{acosta2022validatingroboticssimulatorsrealworld, aljalbout2025reality}, benchmarking against a slower, more accurate physics simulator \cite{pmlr-v211-guo23b, 7139807, 10.1145/2994258.2994272}, or measuring task success rates in learning pipelines \cite{kaup2024reviewphysicsenginesreinforcement}. 
Our approach is most similar to \cite{10.1145/1321261.1321312}, which designs tests with analytical reference solutions. However, \cite{10.1145/1321261.1321312} focuses on simple scenarios such as a box on an incline, a sphere dropped under gravity, and bounce height under restitution. In contrast, our benchmark also focuses on tests that study manipulation behaviors involving force, torque, and contact interactions on both simple and complex object geometries.
We hope this benchmark encourages further research on the evaluation of physical fidelity in simulators, an aspect that remains relatively underexplored despite its significance.
\section{Methods}
\label{sec:method}

Our method builds on top of the unified particle-based solver based on Position-Based Dynamics (PBD) \cite{10.1145/2601097.2601152,10.1145/2994258.2994272, 10.2312/egt.20171034}.
Our novel method, PBD-R, yields a physically accurate PBD formulation that we rigorously evaluate against the analytical solutions provided by our seven-test benchmark.
\subsection{Position-Based Dynamics}
Position-Based Dynamics refers to a family of physics solvers that solve for positions directly using iterative constraint projections \cite{10.2312/egt.20171034}. 
PBD supports a wide range of constraint types to model different physical behaviors, including contact handling, stretching, bending, volume preservation, and others \cite{10.2312/egt.20171034}.
This flexibility enables PBD to model diverse material classes such as rigid bodies, deformable solids, cloth, and fluids. 
To model rigid bodies using particles, three constraints are employed as depicted in \cref{alg:pbd}. (1) Ground Constraint: to prevent particles from penetrating the ground plane. (2) Particle--Particle Contact: To handle contact response between particles that belong to different object groups. (3) Shape Matching Constraint: \cite{10.1145/1073204.1073216} which preserves the relative configuration of particles belonging to the same object so that the object maintains its original shape. 
PBD solves each constraint via a positional correction, which updates particle positions (line \ref{alg:pbd:pos_upd}, \cref{alg:pbd}) and then velocities via $V_{i+1} = \frac{X_{i+1} - X_{\text{init}}}{\Delta t}$ (line \ref{alg:pbd:vel_upd}, \cref{alg:pbd}).
In the following sections, we will describe the modifications (depicted in \textcolor{teal}{teal} in \cref{alg:pbd}) that we applied to the existing PBD formulation for rigid bodies to create a physically accurate formulation, referred to as $\text{PBD-R}$. \\
\textsl{Notations.} Let $X_i$ and $V_i$ denote particle positions and velocities at iteration $i$. For particle $p$, position, velocity, and inverse mass are $x_p$, $v_p$, and $w_p$. Initial particle states are $X_{\text{init}}$, $V_{\text{init}}$, with per-particle values $x_{\text{init}}$ and $v_{\text{init}}$
\begin{algorithm}
\footnotesize
\caption{\footnotesize $\text{PBD-R}$: Position Based Dynamics For Rigid Bodies} \label{alg:pbd}
\textcolor{teal}{Teal text indicates modifications introduced in this work.}
\begin{algorithmic}[1]
\ForAll{particles $p$} $\texttt{ \textbackslash\textbackslash integration step}$ \label{alg:pbd:integration}
    \State $\tilde{v_p} \leftarrow v_{{init}_p} + \Delta t \, w_p f_{\text{ext}}(x_{p})$
    \State $\tilde{x}_{p} \leftarrow x_{{init}_p} + \Delta t \, \tilde{v_p}$
\EndFor
\State $i = 0$, $V_i \leftarrow \tilde{V}$,  $X_i \leftarrow \tilde{X}$
\While{$i < \texttt{solverIterations}$}
    \State $\Delta X_{PG} \leftarrow \texttt{ParticleGroundContact()}$
    \State $\Delta X_{PP} \leftarrow \texttt{ParticleParticleContact()}$
    \State $\Delta X_{SM} \leftarrow \texttt{ShapeMatching()}$ \label{alg:pbd:sm}
    \State $X_{i+1} \leftarrow X_i + (\Delta X_{PG} + \Delta X_{PP} + \Delta X_{SM})$ \label{alg:pbd:pos_upd}
    \State $V_{i+1} \leftarrow \textcolor{teal}{\texttt{Update Velocity()}}$ \label{alg:pbd:vel_upd}
    \State \textcolor{teal}{$X_{i+1}, V_{i+1} \leftarrow \texttt{Enforce Momentum()}$}  
\EndWhile
\end{algorithmic}
\end{algorithm}
\vspace{-3em}
\subsection{Velocity Update Modification}
\label{subsec:vel_updt_modification}
In this section we describe how the existing velocity update $V_{i+1} = \frac{X_{i+1} - X_{init}}{\Delta t}$ (line \ref{alg:pbd:vel_upd}, \cref{alg:pbd}) injects velocity error at every step.
Imagine a simple scenario where particles are receiving a constant force $F$. After the integration step (line \ref{alg:pbd:integration}, \cref{alg:pbd}), the velocity and position of all particles will be $\tilde{V} = V_{init} + a\Delta t$ and $\tilde{X} = X_{init} + V\Delta t$. Now, 
assume all the constraints create a $\Delta X = 0$. This must result in $X_{i}$ and $V_{i}$ to match $\tilde{X}$ and $\tilde{V}$ from the integration step.
In the position and velocity update (lines \ref{alg:pbd:pos_upd}-\ref{alg:pbd:vel_upd}, \cref{alg:pbd})
 We see that $X_{i+1} = X_{i} + 0$,  where $X_{i} = \tilde{X}$, and $V_{i+1} = \frac{X_{i+1} - X_{init}}{\Delta t}$. 
 Analytically, replacing $X_{i+1}$ with $\tilde{X}$, $V_{i+1}$ yields exactly $\tilde{V}$. In finite-precision arithmetic, however, evaluating this large difference over a small $\Delta t$ introduces floating-point errors. Over time, this results in an inaccurate, drifting accumulation of velocity. We address this issue by computing the velocity incrementally as $\tilde{V} + \frac{\Delta X}{\Delta t}$, which is algebraically equivalent to the original PBD update but numerically stable:
 \begin{align}
V_{i+1} 
&= \frac{X_{i+1} - X_{init}}{\Delta t}
= \frac{X_i + \Delta X - X_{init}}{\Delta t} \\
&= \frac{X_i - X_{init}}{\Delta t} + \frac{\Delta X}{\Delta t}
= \tilde{V} + \frac{\Delta X}{\Delta t}.
\end{align}
This decouples velocity from the current position, making both position and velocity depend only on $\Delta X$, limiting accuracy of both by the precision of $\Delta X$.

\subsection{Novel Momentum Conservation Constraint}
\label{subsec:momentum_conservation}
In this section, we describe how the shape matching constraint (line \ref{alg:pbd:sm}, \cref{alg:pbd}) injects linear and angular momentum error, and how we addressed it with our novel constraint. 
Shape matching enforces that all particles belonging to the same body must preserve their relative positions, thereby maintaining the object’s shape. 
It 
achieves this by computing a mass-weighted positional correction that pulls each particle toward a goal position defined by the least-squares optimal transformation from the rest pose to the deformed pose \cite{10.1145/1073204.1073216}. Because the alignment is anchored at the center of mass and weighted by particle mass, momentum must be conserved. In practice, however, finite machine precision leads to the gradual accumulation of numerical drift when performing the shape matching operation, which degrades momentum conservation over time. 

To address this, we introduce a novel constraint that enforces momentum conservation. We compute the linear and angular momentum before and after shape matching, then apply position and velocity corrections so that the post-correction momentum matches the pre-shape-matching momentum values. 
Let $X_{\text{bsm}}, V_{\text{bsm}}$ denote particle positions and velocities before shape matching, and $X_{\text{asm}}, V_{\text{asm}}$ those after shape matching.
We define the linear and angular momentum of a body of particles as
$P = \sum_{p} m_p \, v_p$ and $L = \sum_{p} (x_p - x_{\text{com}}) \times m_p \, v_p$ where $x_{\text{com}} = \frac{1}{M}\sum_p m_p x_p$ is the center of mass
and $M = \sum_p m_p$ is the total mass. 
We denote the linear momentum before and after shape matching as $P_{\text{bsm}}$ and $P_{\text{asm}}$, respectively.

\textsl{Linear Momentum Correction.}
Since the particles belong to a rigid body, we want to calculate a uniform correction for all particles to preserve the body's rigidity. Thus, our goal is to calculate a $\Delta v$ such that
$\sum_p m_p (v_{\text{asm}_p} + \Delta v) = P_{\text{bsm}}.$ 
Solving gives $\Delta v = \frac{P_{\text{bsm}} - P_{\text{asm}}}{M}.$ Applying this correction uniformly to all particles, conserves linear momentum:
\begin{equation}
    v_p \leftarrow v_{\text{asm}_p} + \Delta v, \quad
    x_p \leftarrow x_{\text{asm}_p} + \Delta v \, \Delta t.
    \label{eq:lin_momentum}
\end{equation}

\textsl{Angular Momentum Correction.}
Our goal is to calculate a uniform angular velocity correction \(\omega\) such that the angular momentum of the body matches the value before shape matching was applied:
$\sum_p m_p\, r_p \times \big(v_{\text{asm}_p} - \omega \times r_p\big)
= L_{\text{bsm}}$
where $r_p = x_p - x_{\text{com}}$.
To solve for $\omega$, we first do
$L_{\text{asm}} - L_{\text{bsm}}
= \sum_p m_p\, r_p \times (\omega \times r_p)$
Using the vector triple product identity, we get $L_{\text{asm}} - L_{\text{bsm}} = I\,\omega$ where $I$ is the inertia tensor about the center of mass. The inertia tensor is calculated using $I = \sum_p m_p \left( \lVert r_p \rVert^2 I_3 - r_p r_p^{\top} \right)$ where $I_3$ is the identity matrix. 
Therefore, the angular velocity correction is
$\omega = I^{-1}\Delta L.$
This is applied uniformly to all particles:
\begin{equation}
v_p \leftarrow v_{\text{asm}_p} - \omega \times r_p,
\qquad
x_p \leftarrow x_{\text{asm}_p} - (\omega \times r_p)\,\Delta t
\label{eq:ang_momentum}
\end{equation}
Thus, with \cref{eq:lin_momentum} and \cref{eq:ang_momentum} together, we enforce the conservation of momentum. 

\section{Experiments}
\label{sec:experiments}

\begin{figure*}
\centering
\includegraphics[width=\textwidth]{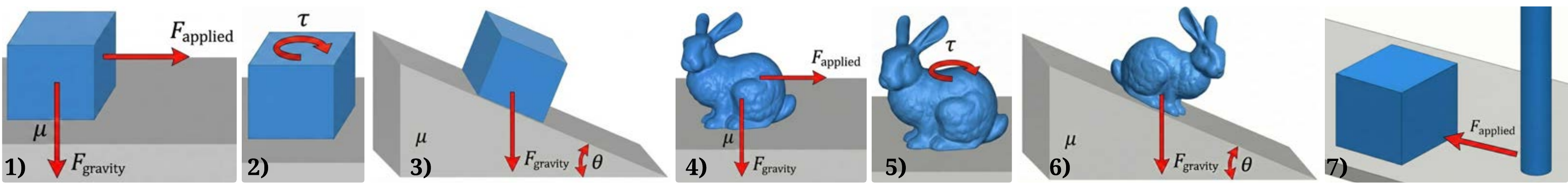}
\caption{%
    We provide this benchmark of physics tests and their analytical solutions to evaluate the physical accuracy of solvers. 
    From left to right: pushed box (Test~1), box with torque (Test~2), box on slope (Test~3),
    pushed bunny (Test~4), bunny with torque (Test~5), bunny on slope (Test~6),
    and rod pushing a box (Test~7). 
}
\label{fig:tests}
\end{figure*}
\begin{table*}[t]
\centering
\caption{%
    Physics test suite. The tests cover four distinct rigid-body behaviors across different objects result in in seven different tests. $F$: applied force magnitude, $\tau$: applied torque magnitude, $M$: total mass, $\mu$: kinetic friction coefficient, $\theta$: slope angle, $\lambda$: principal moment about the rotation axis, $I_{zz}$: moment of inertia about the $z$-axis. CoM: center of mass. Test configurations are illustrated in \cref{fig:tests}.
}
\label{tab:tests}
\small
\setlength{\tabcolsep}{4pt}
\begin{tabularx}{\textwidth}{c l l X}
\hline
\textbf{Test} & \textbf{Name} & \textbf{Analytical Solution} & \textbf{What is tested} \\
\hline
1 & Pushed box        & $x(t) = \frac{1}{2}\frac{(F - \mu M g)}{M}\,t^2$                 & Translation accuracy with a constant force applied at CoM with ground friction \\[6pt]
2 & Box with torque   & $\theta(t) = \frac{1}{2}\frac{\tau}{\lambda}\,t^2$               & Rotation accuracy with a constant torque applied at CoM for a shape with isotropic inertia \\[6pt]
3 & Box on slope      & $x(t) = \frac{1}{2}(g\sin\theta - \mu g\cos\theta)\,t^2$         & Accuracy in resolution of gravity, normal force, and friction on an incline \\[6pt]
4 & Pushed bunny      & $x(t) = \frac{1}{2}\frac{(F - \mu M g)}{M}\,t^2$ & Same as (1) but on a asymmetric mesh \\[6pt]
5 & Bunny with torque & $\theta_z(t) = \frac{1}{2}\frac{\tau_z}{I_{zz}}\,t^2$            & Same as (2) but on an asymmetric mesh with anisotropic inertia \\[6pt]
6 & Bunny on slope    & $x(t) = \frac{1}{2}(g\sin\theta - \mu g\cos\theta)\,t^2$         & Same as (3) on an asymmetric mesh. \\[6pt]
7 & Rod pushing a box & $x(t),\;\theta(t)$ described in \cref{sec:test-design}                                & Accuracy of coupled translation and rotation from force at surface.\\
\hline
\end{tabularx}
\end{table*}

\begin{table}[t]
\centering
\caption{Default parameters for all physics tests unless noted.}
\label{tab:default_params}
\small
\setlength{\tabcolsep}{6pt}
\begin{tabular}{l r @{\;}c@{\;} l}
\hline
\textbf{Parameter} & \multicolumn{3}{c}{\textbf{Value}} \\
\hline
Box mass & $m$ & $=$ & $4\,\mathrm{kg}$ \\
Bunny mass  & $m$ & $=$ & $2.18\,\mathrm{kg}$ \\
Friction coef. in Tests (1), (3), (4), (6), (7) & $\mu$ & $=$ & $0.4$ \\
Friction coef. for Tests (2) and (5) & $\mu$ & $=$ & $0$ \\
Force in Test (1) & $F$ & $=$ & $17\,\mathrm{N}$ \\
Force in Test (4) & $F$ & $=$ & $10\,\mathrm{N}$ \\
Torque in Tests (2) and (5) & $\tau$ & $=$ & $0.01\,\mathrm{N\,m}$ \\
Slope angle in Tests (3) and (6) & $\theta$ & $=$ & $\pi/8$ \\
Rod mass in Test (7) & $m$ & $=$ & $2\,\mathrm{kg}$ \\
Rod force in Test (7) & $F$ & $=$ & $0.1\,\mathrm{N}$ \\
\hline
\end{tabular}
\end{table}
Simulators are routinely validated on task-level metrics such as training reward or sim-to-real transfer success, but these conflate solver accuracy with errors originating from real-world factors and policy design choices. We argue that physical accuracy must be evaluated by isolating rigid-body phenomena and measuring deviation from known analytical solutions. This makes the solver error observable, quantifiable, and independent of any
downstream task. 

\subsection{Physics Tests Benchmark}
\label{sec:test-design}
We present a benchmark covering four distinct rigid-body behaviors, each admitting a closed-form analytical solution: (1) pure translation under a constant force applied through the center of mass, (2) pure rotation under a torque about the center of mass, (3) contact force response involving gravity, normal, and frictional forces on an incline, and (4) coupled translation and rotation arising from contact forces.
To examine the effect of object geometry, we evaluate these behaviors on simple and complex shape geometries. For the first three behaviors, the analytical solutions are object-agnostic, and we evaluate these behaviors on both a cube (simple, symmetric geometry) and the Stanford Bunny (complex, asymmetric mesh). The fourth behavior is specific to objects that maintain single-point contact; therefore, we evaluate it using a rod and a box. Altogether, this yields seven tests depicted in ~\cref{fig:tests}.


\cref{tab:tests} presents the seven test suite. All reference solutions assume ground contact is maintained throughout the trajectory and that the normal degree of freedom is constrained by the ground plane. 
Tests 1–3 use a box geometry to test: pure CoM translation under a horizontal force with kinetic friction (Test 1), pure rotation under a constant torque (Test 2), and combined gravitational, normal, and frictional forces on an inclined plane (Test 3). Tests 4–6 repeat the same three scenarios on the Stanford Bunny, a complex asymmetric mesh. The reference solutions for these six tests are straightforward and presented in \cref{tab:tests}.
We note that to study the pure rotation behavior (Tests 2 and 5), zero friction is assumed. To maintain the object-agnostic property of the rotation behavior tests, we have to assume a zero friction coefficient. Deriving a closed-form rotational friction term requires approximating the contact patch as a disk, which introduces geometric error for non-circular contact patches and yields only an approximate analytical solution.

Test 7, \textsl{Rod Pushing a Box}, evaluates coupled translation and rotation under surface contact. The analytical reference follows the quasi-static pushing model of Lynch and Mason~\cite{lynch1996stable}, which determines the planar velocity of the box $(v_x, v_y, \omega)^T$ from the current configuration, contact geometry, and support friction distribution. Integrating this model forward at fine resolution produces reference trajectories for position $\mathbf{x}(t)$ and orientation $\theta(t)$.

\subsection{Evaluation Setup}
\label{subsec:eval_setup}
We perform a series of analyses with three objectives. Our first objective is to compare PBD\nobreakdash-R against PBD and MuJoCo (\cref{subsec:baseline_comparison}). Our second objective is to study how varying the number of particles affects the PBD\nobreakdash-R output (\cref{subsec:resolution}). Our last objective is to perform an ablation study to understand how PBD\nobreakdash-R addresses the physical accuracy gap compared to PBD (\cref{subsec:ablation}).
We first describe our implementation details and metrics in the following sections and present the results in \cref{sec:results}.

\textsl{Implementation Details.}
We implemented PBD\nobreakdash-R and PBD in the Newton simulator \cite{The_Newton_Contributors_Newton_GPU-accelerated_physics_2025} using Warp \cite{Macklin_Warp_A_High-performance_2022}\footnote{Code available at \href{https://github.com/HIRO-group/newton}{https://github.com/HIRO-group/newton}.}. For MuJoCo, we use the existing GPU implementation developed in Newton. Machine precision is 32-bit floating-point. 
All solvers run at $100\,\text{Hz}$ with 10 substeps ($\Delta t = 1\,\text{ms}$). PBD and PBD\nobreakdash-R use 10 solver iterations. MuJoCo is run for 50 iterations because we observed that it does not converge within fewer iterations. MuJoCo is run with \texttt{njmax = 50} and an elliptic cone friction model, corresponding to the configuration that performed best in our experiments.
All seven tests are run for 1000 frames ($10\,\mathrm{s}$ total at $100\,\mathrm{Hz}$).
We use this duration to capture long-horizon error accumulation that may not be apparent over shorter rollouts. Each test is averaged over three runs with different seeds.

\textsl{Physical Parameters.}
\cref{tab:default_params} summarizes the default parameters used in most experiments. Of note, applied forces, $F$, in Tests (1) and (4) are set to the minimum force that will cause the objects to move, given the friction. In Test (7), the rod applies a force of $F=0.1\,\mathrm{N}$, which is sufficiently small to satisfy the quasi-static assumption underlying the analytical model in \cite{lynch1996stable}.
Regarding the coefficient of friction $\mu$, for all tests except (2) and (5), we set $\mu = 0.4$, corresponding to the ceramic--wood friction reported in \cite{5649517}, representative of real-world manipulation setting \cite{li2024behavior1khumancenteredembodiedai}. For tests (2) and (5), we set $\mu = 0$ because the analytical solution for rotation with friction, where the contact patch is not a disk, is not accurate.
These values serve as default parameters in most experiments. In \cref{subsubsec:sweep_across_mu_F}, we also evaluate robustness to physical parameters by performing a sweep over the coefficient of friction $\mu$ and the applied force $F$. 

\textsl{Shape Representations.} 
We represent objects differently across simulators. In PBD and PBD\nobreakdash-R, objects are defined using particles, whereas in MuJoCo, objects are defined using meshes or geometric primitives. In MuJoCo, the box and rod are modeled as box and cylinder primitives, and the bunny is represented as a triangular mesh.
For PBD and PBD\nobreakdash-R, we convert these geometries into sets of spheres. \cref{fig:first_page} illustrates the resulting sphere-packed representation.
For primitive shapes, we use a deterministic sphere-packing procedure. The box is packed with uniformly spaced, non-overlapping spheres, placing four spheres per axis for a total of 16 spheres. The rod is represented by a single-sphere-wide chain along its length. This ensures approximately one contact point with the box at any given time, following the quasi-static analytical model in \cite{lynch1996stable}.
For mesh geometries, we use a simple sphere-packing algorithm. Given a sphere radius, candidate centers are sampled on a uniform 3D grid over the mesh bounding box, and only those whose centers lie inside the mesh are retained. This produces a non-overlapping set of equal-radius spheres approximating the object's volume. For the bunny, we use a sphere radius of $0.005$, yielding 2,175 spheres, which provides a good geometric approximation of the shape.
These sphere configurations are used in most experiments. In \cref{subsec:resolution}, we further study how varying the number of spheres affects solver accuracy.
\begin{figure*}
\centering
\includegraphics[width=\columnwidth]{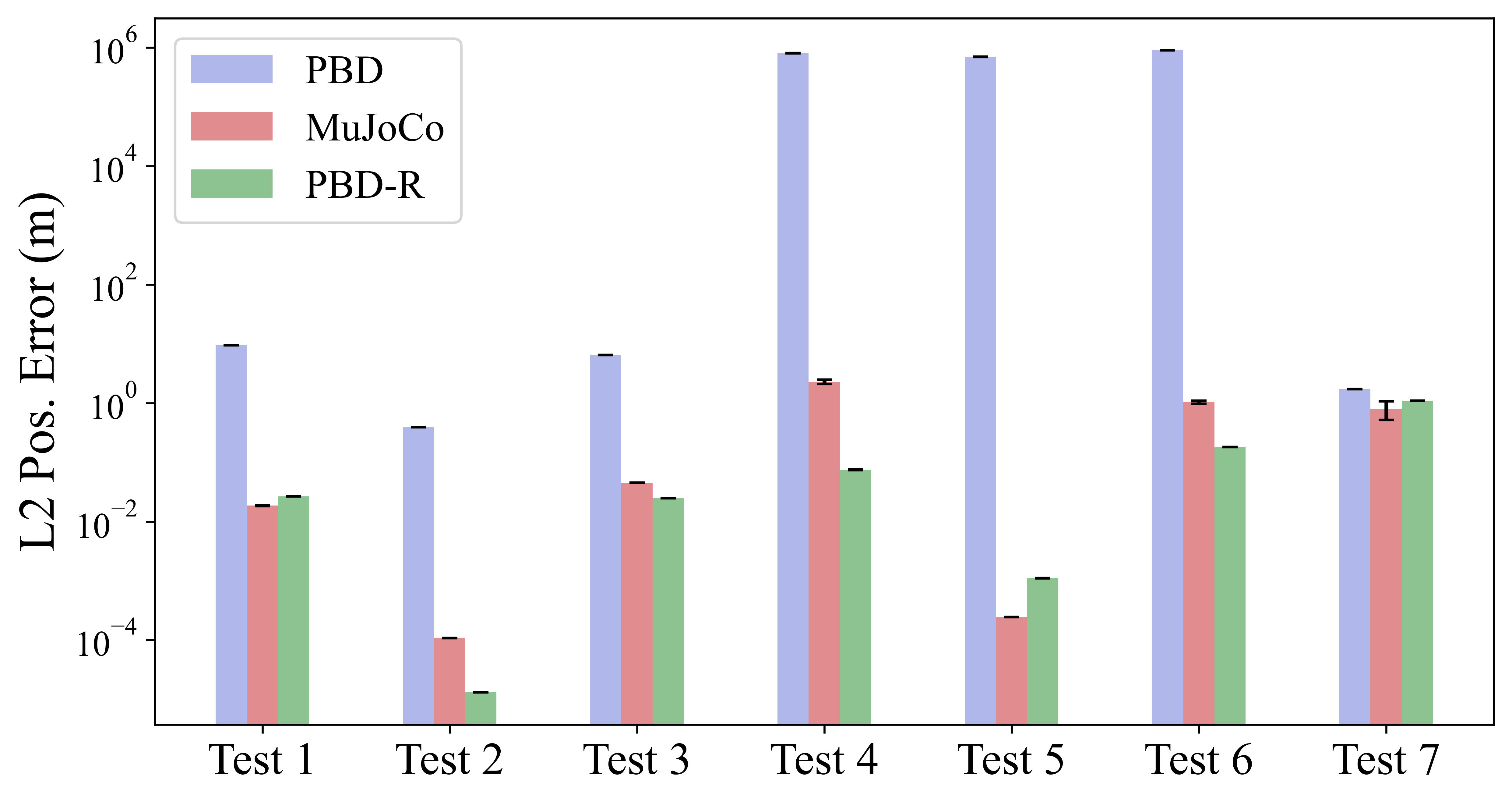}
\includegraphics[width=\columnwidth]{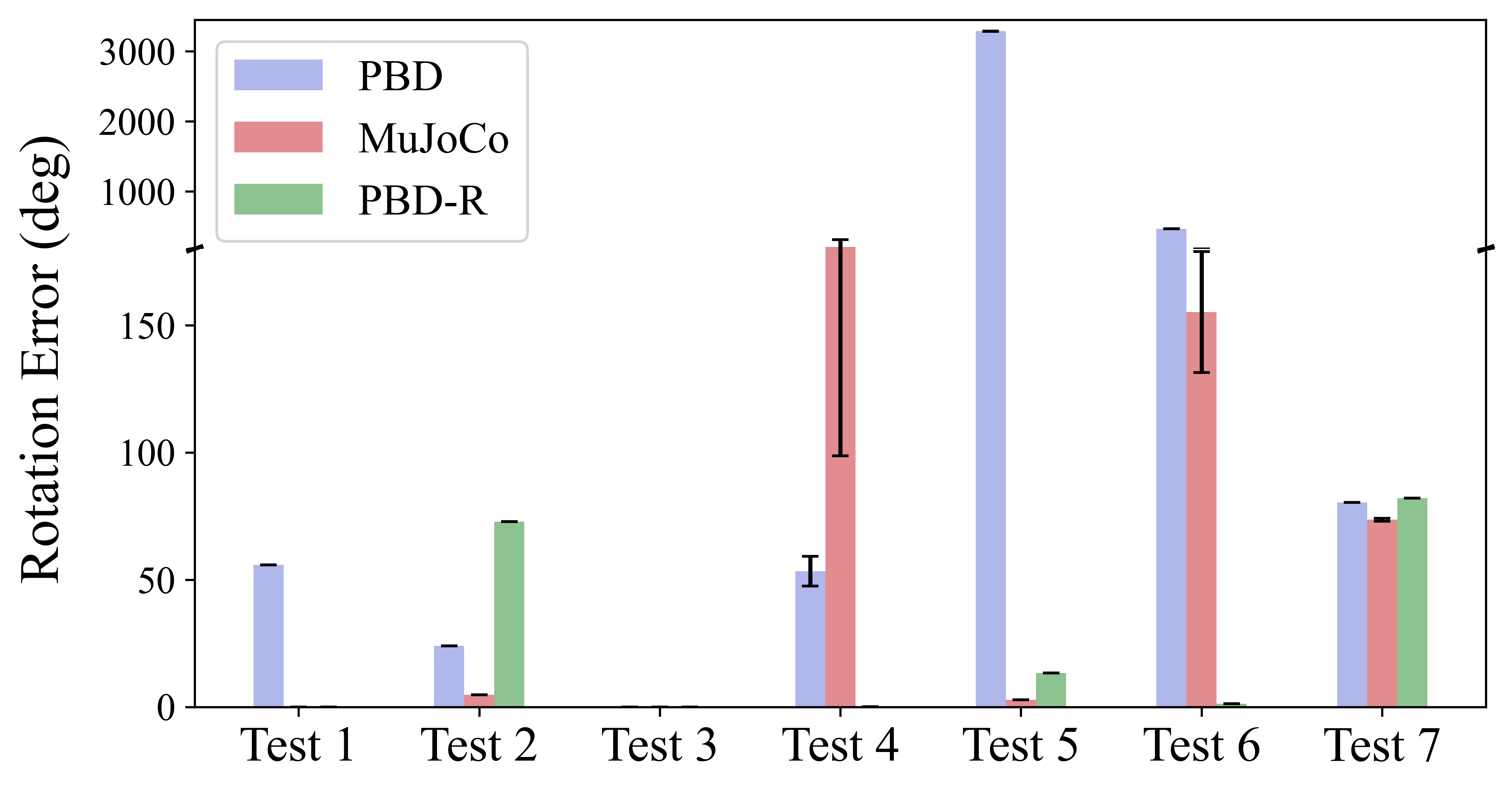}
\caption{Position $\ell_2$ error (left) and rotation error (right) across all seven benchmark tests, averaged over three seeds with standard deviation error bars; rotation errors exceeding $360^\circ$ indicate the simulated orientation has drifted by more than one full revolution from the reference. PBD\nobreakdash-R (ours) consistently outperforms standard PBD and performs competitively with MuJoCo across both metrics.
}
\label{fig:all_exp}
\vspace{-1em}
\end{figure*}

\section{Results \& Discussion}
\label{sec:results}

\begin{figure}
    \centering
    \includegraphics[width=0.7\columnwidth]{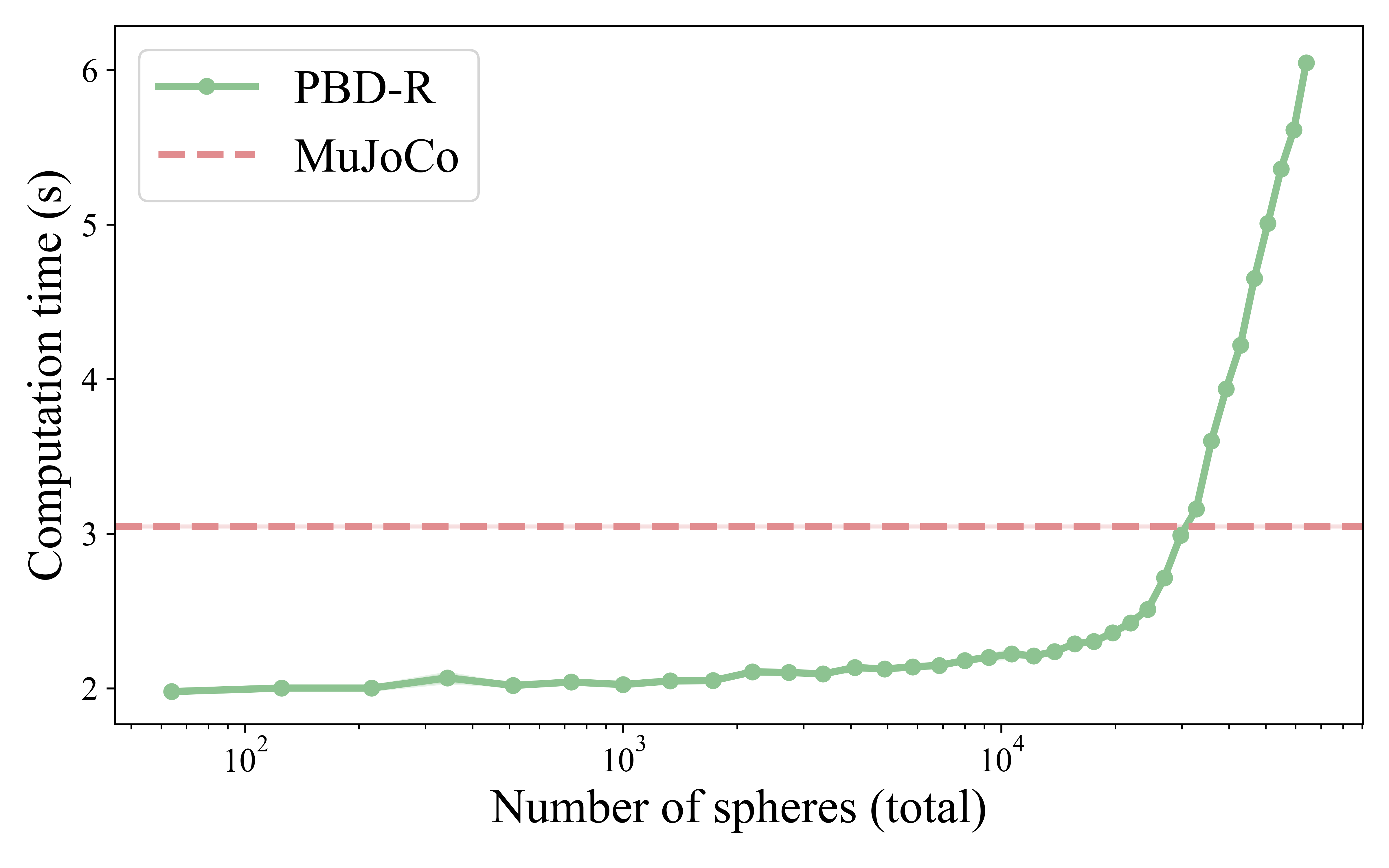}
    \caption{PBD\nobreakdash-R computation time over varying number of spheres compared to MuJoCo over 100 frames of \textsl{Pushed Box} example.}
    \label{fig:computation_times}
\vspace{-1.5em}
\end{figure}

Following the evaluation setup and motivations described in \cref{subsec:eval_setup}, we present our results here.

\subsection{Does PBD-R Help Address the Physical Accuracy Gap?}
\label{subsec:baseline_comparison}
\subsubsection{Comparison Across All Seven Tests}
\label{subsubsec:7_tests_comparison}
\cref{fig:all_exp} compares the final position and rotation errors at the end of the trajectory across the seven benchmark tests. Overall, \cref{fig:all_exp} shows that PBD-R consistently outperforms PBD and achieves accuracy competitive with MuJoCo.
PBD achieves the lowest accuracy among the solvers with position and rotation errors as large as $10^6$ and 3000 degrees, respectively. The large errors are due to drift. Because PBD loses precision during its velocity update and does not conserve momentum, ghost forces accumulate over time, leading to increasing deviations from the expected trajectory. This effect is qualitatively observed in the simulation for Test 7, \textsl{Rod Pushing a Box}. When the rod pushes the box, after a few seconds, the box drifts away and fails to maintain contact with the rod, unlike in the other solvers. Additionally, this drift effect becomes more pronounced as the number of particles increases. Since the bunny is represented using 2175 spheres, compared to 64 spheres for the box, PBD diverges significantly in the tests involving the bunny shape (Tests~4--6). In these tests, position errors grow to the order of $10^6$ m and rotation errors as big as 3000 degrees. 

\begin{figure}
\centering
\includegraphics[width=\columnwidth]{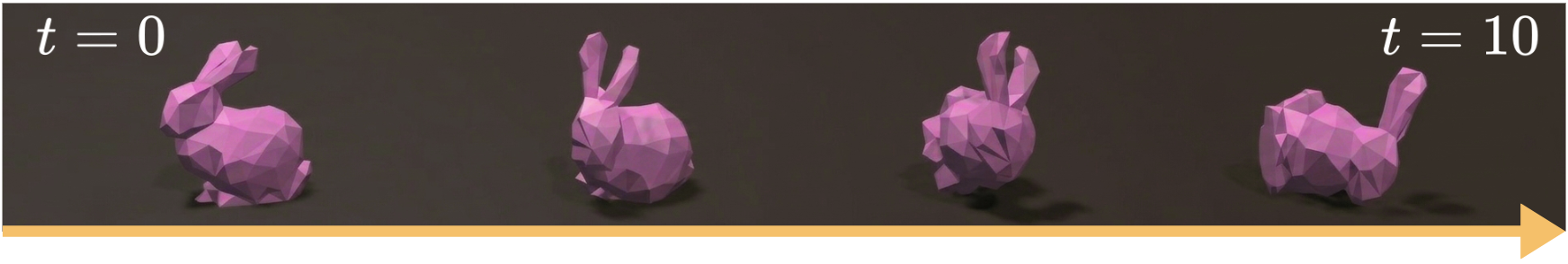}
\caption{The bunny starts tumbling in the \textsl{Pushed Bunny} experiment (Test 4) using MuJoCo.}
\label{fig:bunny_tumbling}
\vspace{-1.5em}
\end{figure}

PBD-R and MuJoCo achieve similar positional accuracy across all tests. Regarding rotation accuracy, Tests~4 and~6 (\textsl{Pushed Bunny} and \textsl{Bunny on Slope}) show large rotation errors for MuJoCo. When we observe the simulation qualitatively, we see that the bunny begins to tumble in both cases (shown in \cref{fig:bunny_tumbling}). This is because the mesh-based contact model generates unstable contact forces between the bunny and the ground that compound over time. In contrast, PBD-R avoids this instability as the contact points are predictably and evenly distributed along the bottom surface of the object. In Test~2 (\textsl{Rotating Box}), PBD-R's rotation error exceeds both baselines. This is due to inertia approximation error at the default sphere resolution under naive packing, an effect we examine in \cref{subsec:resolution}.
\begin{figure}
    \centering
    \includegraphics[width=0.99\columnwidth]{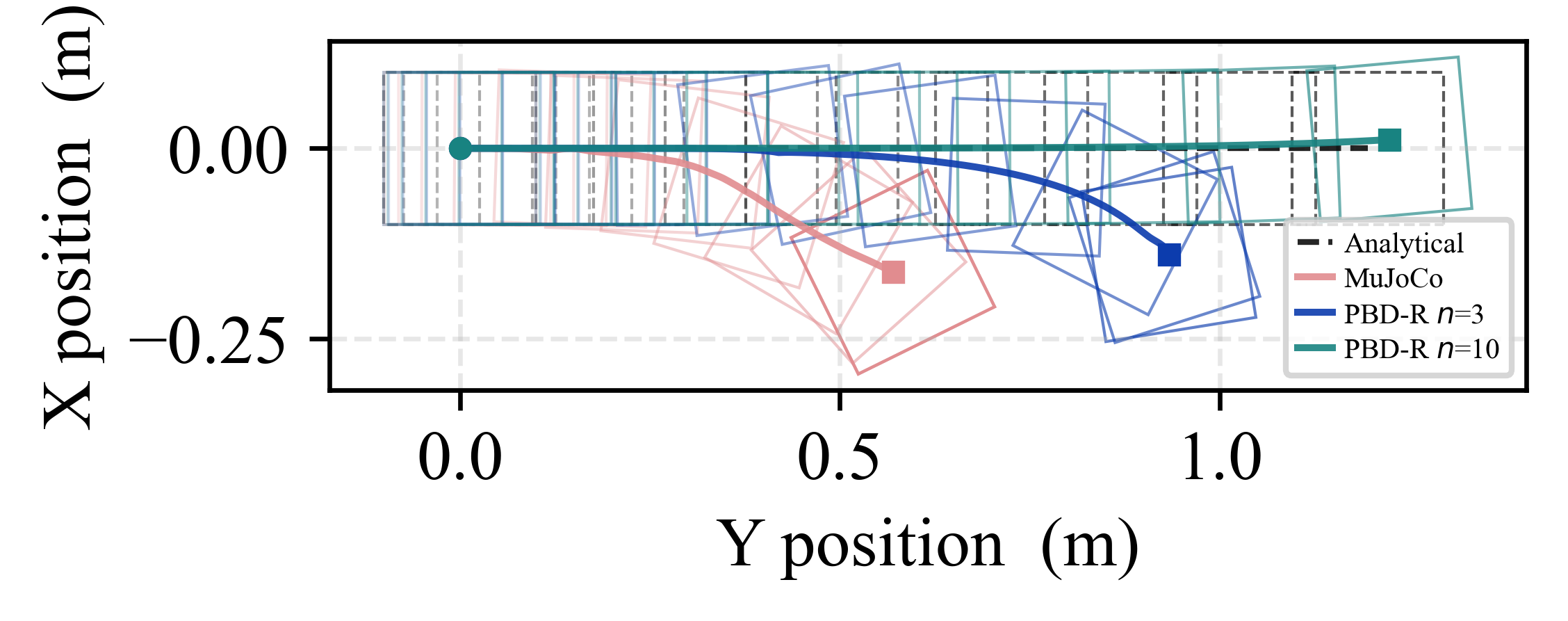}
    \caption{Sample trajectory from Test~7, \textsl{Rod Pushing a Box} (top-down view) with a friction coefficient $\mu=0.4$. Dashed black outlines show the analytical reference. PBD\nobreakdash-R is shown at two resolutions ($n{=}3$ and $n{=}10$ spheres per axis, totaling 27 and 1000 spheres respectively). We use this test to study both the effect of sphere resolution on accuracy and the fidelity of surface contact dynamics under coupled translation and rotation.}
    \label{fig:box_traj}
\vspace{-1.5em}
\end{figure}

\begin{figure}
    \centering
    \includegraphics[width=0.49\columnwidth]{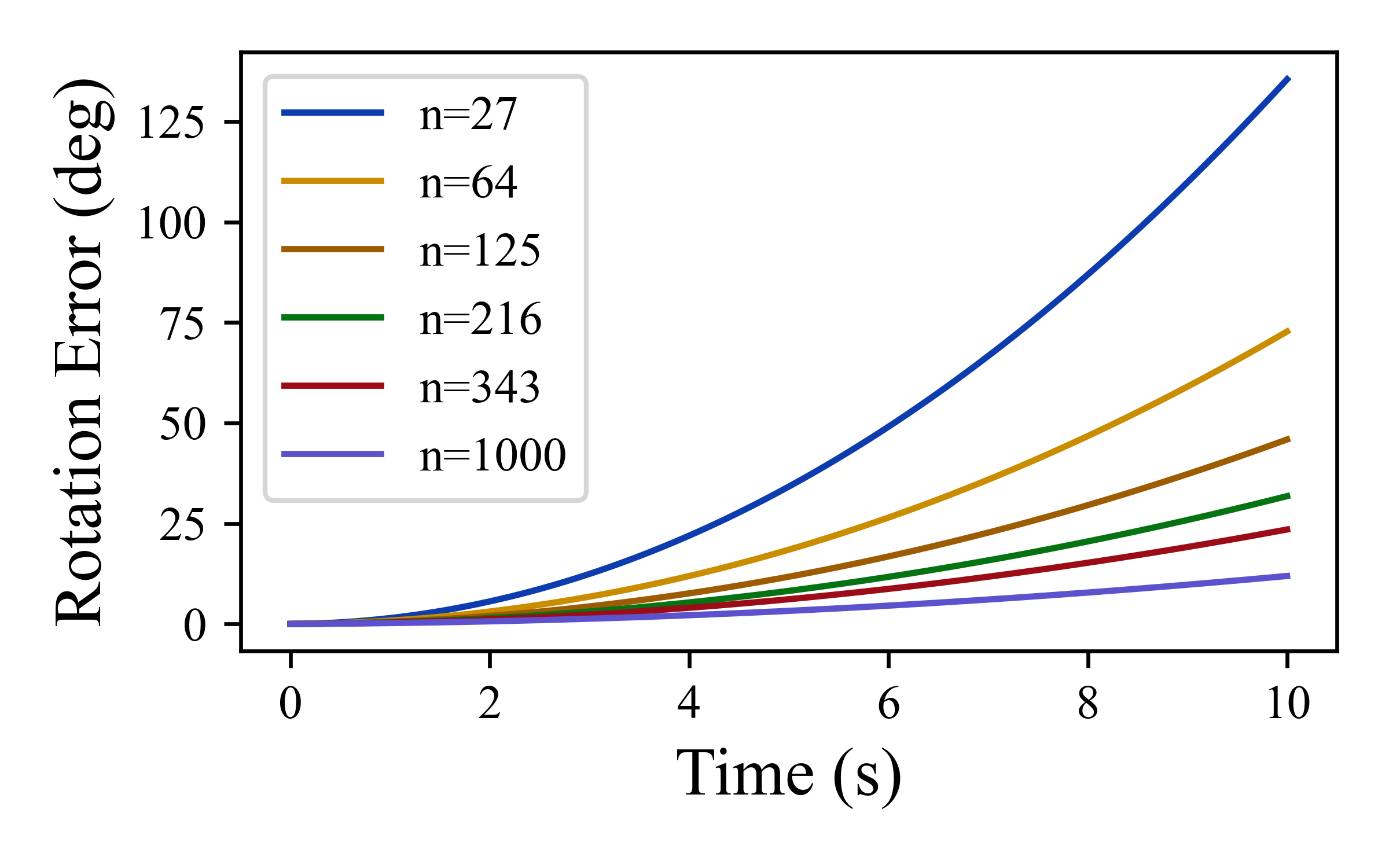}
    \includegraphics[width=0.49\columnwidth]{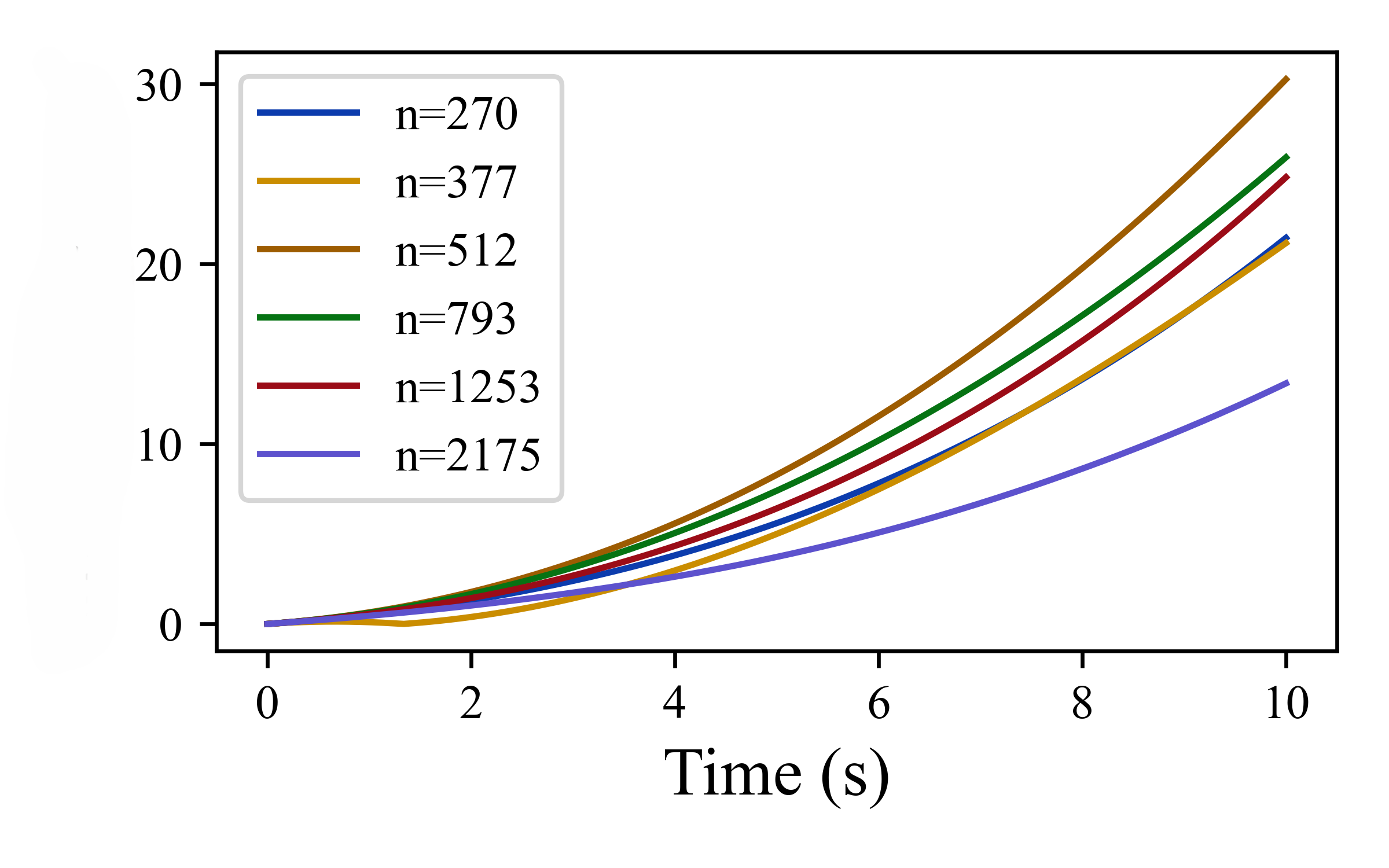}
    \caption{Rotation error during simulation for varying total number of spheres, $n$, in (left) rotate box and (right) rotate bunny examples with constant applied torque. This figure highlights that resolution alone does not determine rotational accuracy; the quality of the volumetric sampling matters as well.
    }
    \label{fig:rot_vs_spheres}
\vspace{-1.5em}
\end{figure}

\subsubsection{Comparison Across Friction Coefficients and Forces}
\label{subsubsec:sweep_across_mu_F}
\begin{figure}
\centering
\includegraphics[width=\columnwidth]{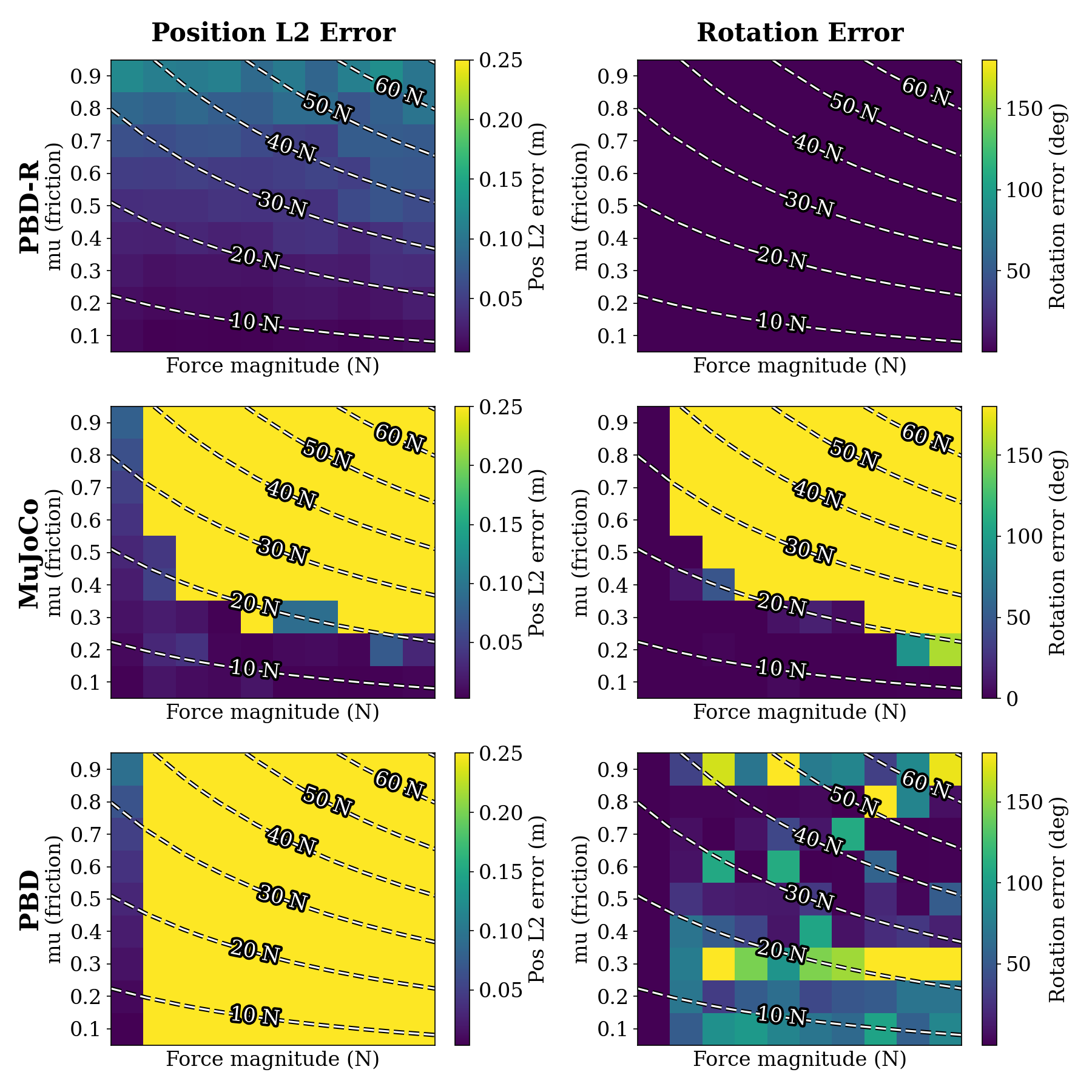}
\caption{Sweep over the friction coefficient and applied force for Test~1 (\textsl{Pushed Box}). The left column plots report the $\ell_2$ position error, while the right column plots show the rotation error. Across the sweep, PBD-R achieves the lowest error, whereas PBD performs the worst. MuJoCo performs well in the regime of low friction and small applied forces, but its accuracy degrades as forces increase.}
\label{fig:sweep_mu_F_box}
\end{figure}

The results in \cref{subsubsec:7_tests_comparison} were obtained at a fixed parameter configuration. To verify that these findings reflect a consistent result across a set of parameters, we sweep over the friction coefficient, $\mu$, and applied force, $F$, for Test~1 (\textsl{Pushed Box}). The results on \cref{fig:sweep_mu_F_box} show the position $\ell_2$ error (left column) and rotation error (right column) for each solver.
PBD-R maintains low error across the entire parameter space, confirming that its accuracy advantage is consistent and not limited to a specific operating condition. Standard PBD performs poorly throughout the parameter sweep, with higher errors at nearly every combination of $\mu$ and $F$. MuJoCo produces higher accuracy at low friction and moderate forces, but degrades noticeably as force increases. Under high force conditions, we observe that the box in MuJoCo begins to tumble, producing large position and rotation errors. This suggests that MuJoCo requires a smaller timestep to remain stable in these regimes, which comes at the cost of increased computational time. The rotation error follows a similar trend to the position error across all three solvers.
Taken together, this sweep establishes that the accuracy observed in \cref{subsubsec:7_tests_comparison} holds broadly: PBD-R is the most robust solver across a range of physical parameters.

\subsubsection{Computation Time Comparison}
\label{subsubsec:comp_time}
\cref{fig:computation_times} compares the wall-clock time required to simulate a box over 100 frames using PBD-R with varying numbers of spheres, while the box primitive in MuJoCo remains unchanged. Experiments use Newton's GPU implementation of MuJoCo on an NVIDIA GeForce RTX 3060.
For up to approximately $10^4$ spheres, PBD-R runs in less than two-thirds of MuJoCo's runtime. Beyond $10^5$ spheres, however, the computational cost increases rapidly. This scaling behavior is a known limitation of the shape matching algorithm \cite{10.1145/1073204.1073216}, whose cost grows superlinearly with particle count.
This computational advantage is especially relevant for robotics, where real-time contact-rich manipulation requires simulators to evaluate candidate actions faster than the control loop frequency \cite{kurtz2026inverse}. 
Moreover, emerging sim-to-real-to-sim pipelines require large-scale simulation rollouts because they iteratively refine the simulator's state by alternating between simulation and real-world experiments \cite{abou-chakra2024physically}.

\subsection{How Does Particle Resolution Affect Accuracy?}
\label{subsec:resolution}
Unlike mesh-based rigid body simulators, where geometric fidelity is largely fixed once the mesh is defined, particle-based representations expose a new degree of freedom: the number of spheres used to discretize an object. We refer to adjusting this quantity as \emph{resolution tuning}. Increasing resolution, i.e., using more and smaller spheres, provides a finer geometric approximation of the object. 
We analyze how this parameter affects PBD\nobreakdash-R 
and identify two primary mechanisms through which resolution influences accuracy:
(1) improved contact modeling due to better geometric approximation.
(2) improved inertial accuracy due to better mass distribution sampling.
These improvements come at a cost, as increasing sphere count raises computational expense (\cref{subsubsec:comp_time}) creating an \emph{accuracy--efficiency} trade-off that is inherent to particle-based simulation.

\begin{figure}
    \centering
    \includegraphics[width=0.8\columnwidth]{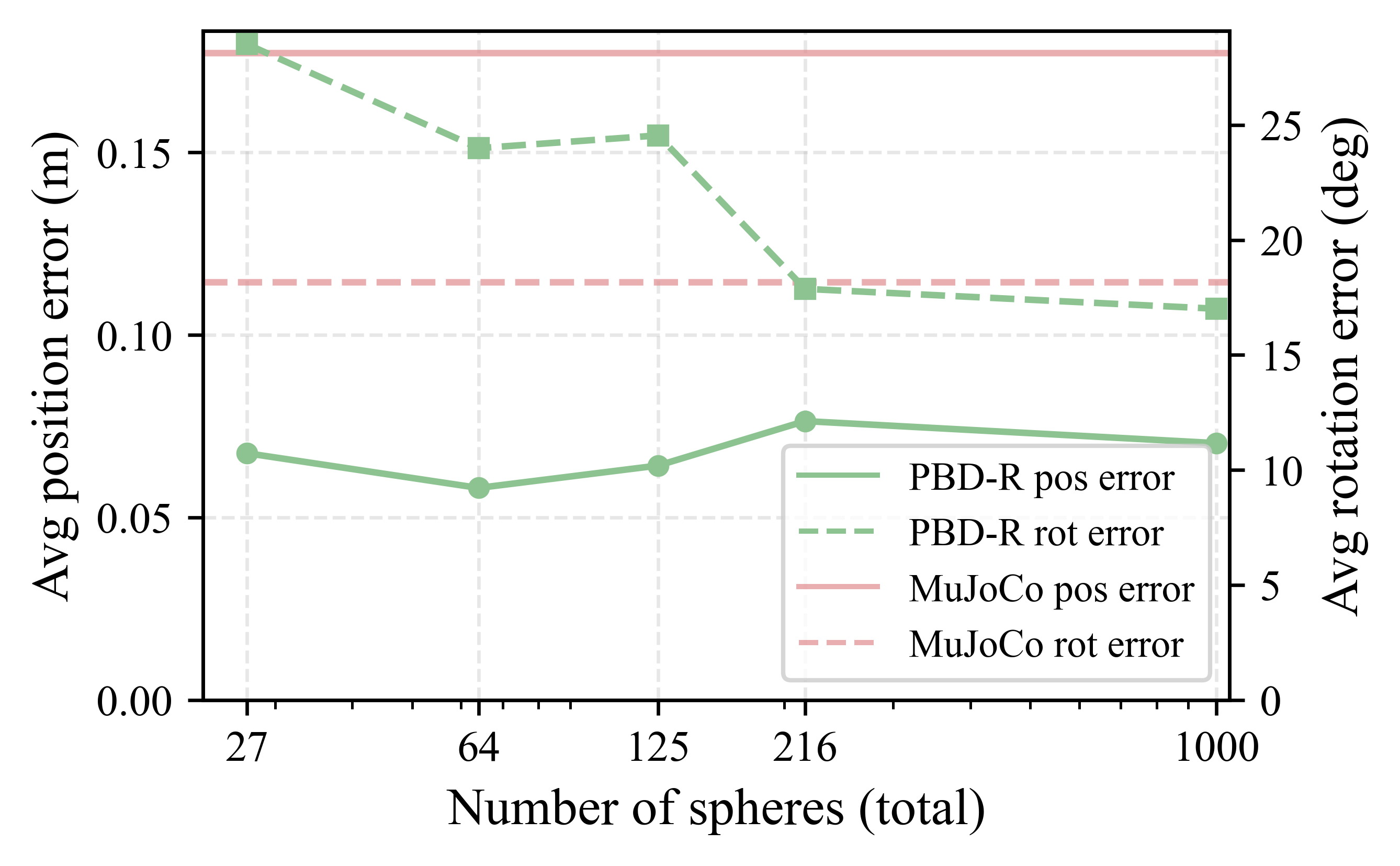}
    \caption{Effect of sphere discretization resolution on trajectory accuracy for Test~7, \textsl{Rod Pushing a Box} across fiction parameters, and push positions. Adding more spheres results in higher accuracy.
    }
    \label{fig:rod_push_err}
\vspace{-1.5em}
\end{figure}

\textsl{(1) Impact of Resolution on Contact Modeling.}
\cref{fig:box_traj} shows a representative trajectory from the \emph{Rod Pushing a Box} 
test, illustrating that the high-resolution discretization of the box ($n{=}10$, 1000 spheres) 
closely tracks the analytical solution, whereas the low-resolution discretization 
($n{=}3$, 27 spheres) exhibits substantial positional and rotational errors due to coarse contact geometry. MuJoCo likewise deviates considerably 
from the analytical trajectory. 
\cref{fig:rod_push_err} generalizes this finding across resolutions, friction parameters, and push positions, showing that PBD\nobreakdash-R's position error decreases monotonically with sphere count, falling below MuJoCo at approximately 216 spheres. 
This improvement arises because finer discretization better approximates the true contact manifold, resulting in more accurate contact force.

\textsl{(2) Impact of Resolution on Inertial Accuracy.}
\cref{fig:rot_vs_spheres} presents rotation error for Tests 2 and 5, \textsl{Rotating Box} and \textsl{Rotating Bunny} respectively, as a function of sphere count. We see that increasing the number of spheres reduces rotational error.
This behavior follows directly from the inertia formulation. For a continuous body, inertia is defined by the volume integral
$\mathbf{I} = \iiint_V \rho(\mathbf{r}) 
\left[ (\mathbf{r} \cdot \mathbf{r}) \mathbf{I}_3 
- \mathbf{r}\mathbf{r}^T \right] dV$
where $\rho(\mathbf{r})$ is the mass density.
In the particle-based model, this integral is approximated by a summation over discrete mass elements.
When the number of spheres is small, the summation provides only a coarse sampling of the mass distribution.
Since angular velocity is $\boldsymbol{\omega} = \mathbf{I}_{\mathrm{com}}^{-1}\mathbf{L}$, errors in inertia $\mathbf{I}_{\mathrm{com}}$ directly affect angular velocity. Because orientation is obtained by integrating $\boldsymbol{\omega}$, these errors accumulate as rotational drift. Increasing the number of spheres improves the inertia approximation, leading to more accurate angular velocity and lower rotational error.
Taking a closer look at \cref{fig:rot_vs_spheres}, the two geometries show different behaviors.
For the box, error decreases consistently with resolution, which follows from its structured discretization: spheres are placed on a uniform grid, so the discrete mass distribution converges smoothly to the true inertia as resolution increases.
The bunny, however, exhibits inconsistent behavior, with lower-resolution configurations ($n=270$, $n=377$) achieving smaller errors than several higher-resolution ones.
This occurs because the bunny is discretized via a naive sphere-packing strategy, retaining only grid-sampled candidates whose centers fall inside the mesh. This procedure does not guarantee uniform volumetric coverage, so certain lower-resolution samplings happen to approximate the true inertia tensor more accurately than higher-resolution ones.
These results highlight that rotational accuracy in PBD\nobreakdash-R depends not only on the number of spheres, but on their spatial distribution. A principled sphere-packing strategy~\cite{nechyporenko2025morphit} that optimizes volumetric coverage can achieve higher fidelity at the same sphere count, offering a more efficient path along the accuracy--efficiency trade-off.
\subsection{Ablation for Velocity Update and Momentum Constraint}
\label{subsec:ablation}
PBD-R introduces two modifications to standard PBD: a corrected velocity update (\cref{subsec:vel_updt_modification}) and a momentum conservation constraint (\cref{subsec:momentum_conservation}). To verify that each is necessary, we remove each component on Test 1 (\textsl{Pushed Box}) and report the resulting degradation in \cref{tab:ablation}. Across all ablations, the errors are initially small (e.g., at $t{=}2$s) but grow substantially by $t{=}10$s, indicating that small numerical violations accumulate over time and lead to large trajectory deviations.
Removing the corrected velocity update allows floating point errors to accumulate in the integration step, causing position error to grow to 1.33\,m by $t{=}10$\,s (310$\times$ increase in mean error) and velocity error to diverge similarly. Removing linear momentum preservation allows small violations introduced during shape matching to accumulate, producing momentum drift that reaches 6.88\,kg${\cdot}$m/s by $t{=}10$\,s and drives position error to 3.95\,m. Removing angular momentum preservation has an analogous effect on rotation, with orientation error reaching 135.85\textdegree{} by $t{=}10$\,s.
Both modifications address distinct sources of numerical drift: the velocity update corrects the integration step, while the momentum constraint corrects shape matching. Removing either one produces physically meaningful degradation (\cref{tab:ablation}), validating their inclusion in PBD-R.
\section{Conclusion}
\label{sec:conclusion}
\begin{table}[t]
\centering
\vspace{1em}
\caption{Ablation study. Each row shows the error when a component is removed from PBD\nobreakdash-R (ours maintains ${\approx}0$ error across all metrics). The $\times$ column shows the ratio of ablated to ours.}
\label{tab:ablation}
\setlength{\tabcolsep}{2.5pt}
\footnotesize
\begin{tabular}{@{}ll rrrr@{}}
\toprule
Ablation & Metric & $t{=}2$s & $t{=}10$s & Mean & $\times$ \\
\midrule
\multirow{2}{*}{Velocity Update}
 & Pos.\ (m)    & 2e-4  & 1.33  & 0.32  & 310$\times$  \\
 & Vel.\ (m/s)  & 9e-3  & 0.21  & 0.18  & 2403$\times$ \\
\midrule
\multirow{2}{*}{Linear Momentum}
 & Pos.\ (m)                      & 0.03  & 3.95  & 0.73  & 703$\times$  \\
 & Lin.\ (kg${\cdot}$m/s)         & 0.12  & 6.88  & 1.58  & 5063$\times$ \\
\midrule
\multirow{2}{*}{Angular Momentum}
 & Rot.\ (deg)                         & 14.99 & 135.85 & 64.74 & 1906$\times$ \\
 & Ang.\ (kg${\cdot}$m$^2$/s)        & 6e-3  & 5e-3   & 6e-3  & 828$\times$  \\
\bottomrule
\end{tabular}
\vspace{-1em}
\end{table}
In this work, we introduce PBD-R, a revised Position-Based Dynamics formulation for particle-based simulation that improves the rigid body physical accuracy of PBD through a corrected velocity update and a novel momentum-conservation constraint. We further introduce a solver-agnostic benchmark of seven physics tests with closed-form analytical reference solutions. Using these tests, we show that PBD-R substantially reduces physical error compared to standard PBD and achieves accuracy competitive with MuJoCo at lower computational cost.
Particle-based representations offer compelling advantages, including a unified representation for diverse materials, fast collision checking, and no need for convex decomposition. However, they also introduce unique modeling challenges. In particular, modeling flat surfaces using particles is inefficient. Moreover, solver accuracy depends on the number of particles and particle placement, making effective sphere-packing strategies critical for both simulation accuracy and computational efficiency. Exploring effective packing methods remains an important direction for future work. Looking forward, we plan to explore dynamic resolution control, where particle density can adapt to task requirements or interaction regions.
Finally, we plan to integrate PBD-R into closed-loop robot control pipelines and evaluate whether improved physical fidelity translates into reliable sim-to-real transfer across manipulation tasks.
\AtNextBibliography{\footnotesize}
\printbibliography

\end{document}